\documentclass[runningheads,a4paper]{llncs}

\usepackage{amssymb}
\usepackage{graphicx}
\usepackage{float}
\usepackage{subfloat}
\usepackage{subcaption}
\usepackage{xcolor}
\usepackage{multirow}
\usepackage{url}
\usepackage{longtable}
\usepackage{todonotes}
\usepackage{tabularx}
\urldef{\mailsa}\path|{alfred.hofmann, ursula.barth, ingrid.haas, frank.holzwarth,|
\urldef{\mailsb}\path|anna.kramer, leonie.kunz, christine.reiss, nicole.sator,|
\urldef{\mailsc}\path|erika.siebert-cole, peter.strasser, lncs}@springer.com|    
\newcommand{\keywords}[1]{\par\addvspace\baselineskip
\noindent\keywordname\enspace\ignorespaces#1}
\begin{document}

\mainmatter  

\title{Near Real-Time Object Recognition for Pepper based on Deep Neural Networks Running on a Backpack}

\titlerunning{Object Recognition for Pepper Robot on a Backpack}

%
%
\author{Esteban Reyes\inst{1}, Cristopher G\'omez\inst{1}, Esteban Norambuena\inst{1} \and Javier Ruiz-del-Solar\inst{1, 2}}
\authorrunning{Esteban Reyes et al.}

\institute{Department of Electrical Engineering, Universidad de Chile
\email{\{esteban.reyes,cristopher.gomez,esteban.norambuena\}@ug.uchile.cl}
\\[0.1cm]
\and
Advanced Mining Technology Center, Universidad de Chile\\
\email{jruizd@ing.uchile.cl}}

\toctitle{Lecture Notes in Computer Science}
\tocauthor{Authors' Instructions}
\maketitle

\begin{abstract}

The main goal of the paper is to provide Pepper with a near real-time object recognition system based on deep neural networks. The proposed system is based on YOLO (\emph{You Only Look Once}), a deep neural network that is able to detect and recognize objects robustly and at a high speed. In addition, considering that YOLO cannot be run in the Pepper's internal computer in near real-time, we propose to use a Backpack for Pepper, which holds a Jetson TK1 card and a battery. By using this card, Pepper is able to robustly detect and recognize objects in images of 320x320 pixels at about 5 frames per second. 

\end{abstract}

\section{Introduction}

Environments where service and social robots operate/live are highly dynamic, in part because of the people living in those environments constantly interact with each other and carry out daily life activities. Therefore, service and social robots require perception systems that are highly robust, but at the same time are able to operate in real-time\footnote{In the service robot domain, we understand as real-time tasks that have a reaction span that looks natural to people, i.e. $\sim$5 Hz.}. Object detection and recognition are some of the vision tasks that require at least near real-time operation.

The commercial Pepper robot \cite{robotics2017pepper} is a social robot used to research on human-robot interaction in real-world human environments (e.g. it is the official platform of the RoboCup@Home Standard Platform League), but its operation is constrained to the fact that it lacks the computational power necessary to run state-of-the-art vision algorithms. 

On the other hand, the uprising of Deep Neural Networks (DNNs) has lead researchers to use them in the development of models that can quickly recognize objects and persons in a robust manner. However, they are very expensive in terms of computational power, so they cannot be run directly on typical service robots. In particular, most of state-of-the-art DNN models cannot run on Pepper's internal computer, a situation that stands as the problem we address in this paper. Therefore, the main goal of this paper is to propose a solution to this problem that considers  two components: First, the selection of YOLO (\emph{You only look Once}) \cite{redmon2016you}, a DNN with real-time detection and recognition of objects and persons, as the most suitable DNN to this task; and second, the development and implementation of an add-on for Pepper, a backpack that permit the attachment of a single board computer onto Pepper, particularly a Nvidia Jetson TK1, which can run YOLO at about 5 FPS when processing images of 320x320 pixels. The use of an external enhancing device, attached to the robot without modifying its structure, comes as an inspiration from similar projects developed for the NAO robot \cite{mattamala2017nao}, where a fully replicable backpack was built.

In this work we provide details on how to reproduce the backpack for Pepper, and on how to install and implement YOLO on it, making the backpack CAD model, hardware, software specifications and an installation guide available for replication. 

The remainder of this paper is structured as follows. In Section 2, we present an overview of the YOLO network. Subsequently, in Section 3 we present the process done to adapt and use YOLO on the Pepper's internal computer and in the Jetson TK1 card. Then, in Section 4 we address the mechanical design of the Pepper Backpack. Finally, in Section 5 we draw some conclusions of this work.

\section{Robust and Fast Object Recognition using YOLO}

\emph{You Only Look Once} (YOLO) \cite{redmon2016you}, is a computer vision system capable of detecting a wide variety of objects in a single image, with an accuracy similar to RetinaNet \cite{lin2017focal}, but with a superior speed of inference when compared to others state of the art systems such as SSD \cite{liu2016ssd}, R-FCN \cite{dai2016r} and FPN FRCN \cite{lin2017feature}. Its speed makes it one of the best-suited systems for real-time object detection needed in systems such as service robots.

One of the main features of YOLO is to treat the detection of objects as a regression problem, where the model is trained to identify bounding boxes (BB), along with probabilities of certainty, in areas that might contain an object. Unlike other systems, where \emph{proposals} or \emph{regions of interest} are generated explicitly as object candidate windows, on which a class inference is executed. This difference gives YOLO advantages in terms of speed, having to process each image only once to perform multiple detections, instead of processing \emph{proposals} individually. Also, the system processes the complete image and not just a region of it, which makes its inferences contemplate the global context of the image, making it less likely to detect background content as an object, which translates into a lower number of false positives in comparison to other systems.

Another distinctive feature of this model is its end-to-end training process, meaning that it has a unique \emph{pipeline} that is trained jointly. Unlike other systems that have different components and need to be trained separately, such as Faster-RCNN \cite{ren2015faster}.

Since the first release of YOLO, there have been 3 major versions of the algorithm, each one aiming to improve accuracy with respect to previous versions. The first version, YOLOv1 \cite{redmon2016you}, achieved 63.4\% mean average precision (mAP) over PASCAL VOC 2007 \cite{pascal-voc-2007}, with an inference speed of 45 FPS. Introduction of a fully-convolutional model \cite{liu2016ssd}, multi-scale training \cite{redmon2017yolo9000}, batch normalization \cite{ioffe2015batch}, BB dimensions priors \cite{redmon2017yolo9000}, among other techniques, raised YOLOv2 \cite{redmon2017yolo9000}, which gets an mAP of 48.1\% on COCO \cite{lin2014microsoft} dataset and 78.6\% mAP on PASCAL VOC 2007, while working at 40 FPS. The latest version of YOLO, YOLOv3 \cite{yolov3}, includes a larger model with 75 convolutional layers that use residual blocks \cite{he2016deep}, prediction of BB across 3 different scales by using a procedure similar to feature pyramid networks \cite{lin2017feature}, among other improvements that result in an mAP of 57.9\% on COCO, at 20 FPS on the same TitanX GPU that all models where tested. Differences among YOLO versions clearly depicts that performance can be sacrificed to gain processing speed, allowing to choose the best-suited version for a given application. Moreover, there are reduced versions of YOLOv1 and YOLOv2 which are even faster. A review of most versions of YOLO is shown in Table \ref{table:YOLOversions}.

\begin{table}[H]
\centering
\caption{Performance of different YOLO versions. Inference speed results (FPS) where obtained when running on a Titan X GPU. Fastest YOLO version is tiny-YOLO v2 at 207 FPS, and most accurate version is YOLOv3 with a 57.9\% mAP on COCO dataset.}
\label{table:YOLOversions}
\begin{tabular}{c|c|c|c|c|c}
\textbf{Model} & \textbf{Input size} & \textbf{Train set} & \textbf{Test set} & \textbf{mAP}    & \textbf{FPS} \\ \hline \hline
YOLOv1         & 448x448             & VOC 2007+2012      & VOC 2007          & 63.4\%          & 45           \\
Fast YOLOv1    & 448x448             & VOC 2007+2012      & VOC 2007          & 52.7\%          & 155          \\
YOLOv2         & 416x416             & VOC 2007+2012      & VOC 2007          & 76.8\%          & 67           \\
tiny-YOLOv2    & 416x416             & VOC 2007+2012      & VOC 2007          & 57.1\%          & \textbf{207} \\ \hline
YOLOv2         & 608x608             & COCO               & COCO              & 48.1\%          & 40           \\
YOLOv3         & 608x608             & COCO               & COCO              & \textbf{57.9\%} & 20          
\end{tabular}
\end{table}

\section{Adapting YOLO to be used with Pepper Robots}

 In spite of YOLO's exceptional performance and speed, it is necessary to take into account that the reported speeds were measured using a platform with a powerful GPU. When running YOLO on a CPU-only system, such as the Pepper's internal computer, the speed of operation decreases considerably. To deal with this situation is that we base our work on the fastest versions of the model (tiny-YOLO), which has only 15 convolutional layers.

\subsection{YOLO for Pepper}

To run YOLO on the Pepper's on-board computer, the Darknet framework \cite{redmon2013darknet} must be first installed. Darknet is written in C, a low-level language, so it is easy to port to different platforms. Thus, compilation for Pepper's computer is straightforward.

To easily integrate YOLO with other modules of Pepper, the \texttt{darknet\_ros} package from ETH Autonomous System Lab  \cite{darknet_ros} is used. This package operates with an older version of Darknet, however, it is compatible with the most recent versions of Darknet. The YOLO ROS package implements all the tools needed to feed the system with a standard ROS stream of images through a defined topic. Moreover, the information of the detected objects is also published through a ROS topic. 

To compile the YOLO ROS package for the Pepper's computer, mild modifications are needed to source and compilation files. A guide to the process can be found in the \emph{Supplementary Material} Section. 

\subsection{YOLO for Jetson TK1}

As an alternative strategy to the Pepper's on-board processing of tiny-YOLO, an external processing unit is used, where the only task that directly involves Pepper is to publish on a ROS topic the images from his camera at a rate that depends on the resolution of the images. The maximum resolution that enables a rate of $\sim 30$Hz is 640x480. Jetson TK1 development card is used for the external processing. We select this unit because it has an integrated 2GB Nvidia GPU, which allows the use of parallel computing platform CUDA \cite{sanders2010cuda}, accelerating the calculation of certain Darknet operations written exclusively for parallel processing in CUDA.

As a first approach, the original version of Darknet is compiled, it is important to note that its compilation with GPU is immediate, but not the use of the cuDNN library \cite{chetlur2014cudnn} of efficient computing for deep learning. This incompatibility raises because of the CUDA version available for the Jetson TK1 is CUDA 6.5, and the library cuDNN from Darknet uses the CUDA 7.0 version, thus processing speeds will be lower.

Afterwards, YOLO ROS is compiled on the Jetson card, for which ROS Indigo was previously installed by following the ROS Jetson guide \cite{jetson_ros}. The YOLO ROS compilation without GPU is straightforward, but to enable its use some modifications need to take place. The changes made to YOLO ROS focuses on the file \texttt{CMakeList.txt}, the scripts in C++, the configuration of the CUDA paths and extensions of some ROS Indigo files. Due to the extensive and tedious modifications made to the code, they will be omitted, but these and the entire installation process in the Jetson TK1 are reviewed in detail in the \emph{Supplementary Material} Section.

\begin{figure}[h]
\centering
\includegraphics[width=1.0\textwidth]{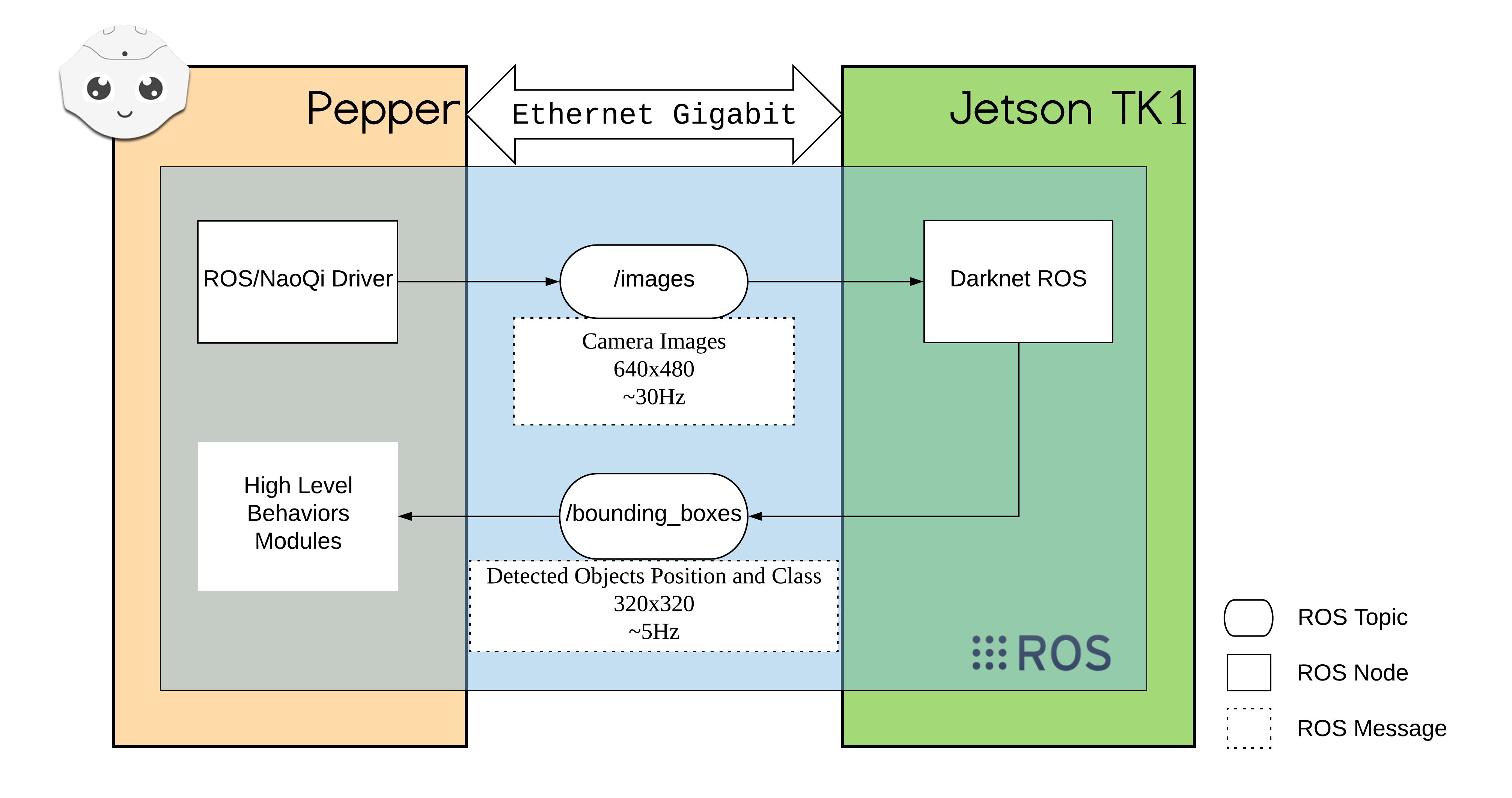}
\caption{Pepper to Jetson TK1 communication diagram. Through the ROS/NaoQi driver, Pepper publishes images from his camera. The Darknet ROS node runs the YOLO network that feeds from the images. The detected objects and their classes are published in a custom ROS message, which can be easily accessed by software modules running in Pepper's internal computer.}
\label{fig:architecture-diagram}
\end{figure}

Finally, a \emph{ROS Network} is configured to connect Pepper to the Jetson through an Ethernet cable. Configuring Pepper as MASTER and Jetson as a HOST. This enables the access of Jetson TK1 card to the topic where images from Pepper's camera are published, then process them with tiny-YOLO to finally publish detection information on another topic that Pepper can subscribe to and use as desired. A diagram that depicts the connections made is shown in Fig. \ref{fig:architecture-diagram}.

\subsection{Results}


The training process of tiny-YOLO considers input images that vary in resolution. In the process, the images are resized every ten training iterations, to resolutions equal to $320 + 32n$, where $n\in \{0,1,...,9\}$ is chosen randomly. By doing this, images from 320x320 pixels up to 608x608 pixels are considered in the training process. This multi-scale property allows to choose the input image resolution at runtime; so the input image size is used as a system parameter to manage the trade-off between inference speed and detection accuracy.  The original tiny-YOLO configuration, that considers 416x416 pixel input images, is first tested on a notebook with a quadcore i5-4210U 1.70GHz CPU, and the inferences speed reached roughly 1 FPS. Then, input images are reduced to 160x160 pixels, allowing detection of big objects at $\sim$6 FPS, while small ones were not detected. Afterwards, we compared tiny-YOLO and tiny-YOLO ROS in the notebook, Pepper, and Jetson at a fixed input size of 160x160, these results are summarized in Table \ref{table:FPS}. 

\begin{table}
\centering
\caption{Comparison between different tiny-YOLO systems and platforms. The inference speed and image size are evaluated.}
\label{table:FPS}
\begin{tabular}{cccc}

\textbf{Model} & \textbf{Computer} & \textbf{\begin{tabular}[c]{@{}c@{}}Input image size\end{tabular}} & \textbf{\begin{tabular}[c]{@{}c@{}}Inference \\ speed [FPS] \end{tabular}} \\ \hline \hline
tiny-YOLO & \begin{tabular}[c]{@{}c@{}}Notebook CPU\\ quadcore i5-4210U 1.70GHz\end{tabular} & 160x160 & $\sim$6 \\ \hline
tiny-YOLO ROS & \begin{tabular}[c]{@{}c@{}}Notebook CPU\\ quadcore i5-4210U 1.70GHz\end{tabular} & 160x160 & $\sim$3 \\ \hline
tiny-YOLO & Pepper & 160x160 & $\sim$0.6 \\ \hline
tiny-YOLO ROS & Pepper & 160x160 & $\sim$0.3 \\ \hline
tiny-YOLO ROS & \begin{tabular}[c]{@{}c@{}}ROS Network\\ Pepper-Jetson\end{tabular} & 160x160 & $\sim$16 \\ 
\end{tabular}
\end{table}

To consider the object detection method as a near real-time system, it must have an inference speed of at least 5 FPS. Table \ref{table:FPS} makes clear that the Pepper's on-board computer cannot be used to run the tiny-YOLO ROS system, so incorporation of the Jetson TK1 unit is needed. In Table \ref{table:FPS} is shown that tiny-YOLO ROS running on Jetson TK1 jointly with Pepper reaches 16 FPS, which gives a margin to enlarge input image size and improve detections accuracy at cost of decreasing inference speed. Experiments on how inference speed decreases when making input image size bigger, are shown in Table \ref{table:FPSvsSize}, where mAP over VOC 2007 test set is calculated to depict that a greater image size means better detection, specifically of small objects. The highest resolution that could be tested was 384x384 because a greater one leaves Jetson TK1 without computational resources.

\begin{table}
\parbox{.485\textwidth}{
\centering
\caption{Speed of tiny-YOLO ROS for Pepper-Jetson TK1 at different input image size. Inference speed decreases as input size increases, but larger images get higher mAP on VOC 2007 test set.}
\label{table:FPSvsSize}
\begin{tabular}{ccc}
\textbf{\begin{tabular}[c]{@{}c@{}}Input image\\size\end{tabular}} & \textbf{\begin{tabular}[c]{@{}c@{}}Inference\\ speed {[}FPS{]}\end{tabular}} & \textbf{mAP} \\ \hline \hline
160x160                                                             & $\sim$15.5                                                          & 24.60\%            \\
224x224                                                             & $\sim$5.9                                                                    & 37.10\%            \\
288x288                                                             & $\sim$5.8                                                                    &44.72\%            \\
\textbf{320x320}                                                             & \textbf{$\sim$4.8}                                                                    & \textbf{47.69\%}            \\
352x352                                                             & $\sim$4.5                                                                    & 50.32\%            \\
384x384                                                             & $\sim$3.6                                                                    & 52.62\%   
\end{tabular}
}
\hfill
\parbox{.485\textwidth}{
\centering
\caption{Class-wise mean average precision (mAP) on VOC 2007 test set for tiny-YOLO ROS in Pepper-Jetson TK1, with an input image size of 320x320.}
\label{table:320classes}
\begin{tabular}{ccccc}
aero & bike & bird & boat & bottle \\ \hline
47   & 60.1 & 39.5 & 30.6 & 15.8  \\
bus  & car  & cat  & chair & cow  \\ \hline
61.2 & 57.8 & 65.6 & 23.9  & 43.1 \\
table & dog  & horse & m bike & person \\ \hline
49    & 59.2 & 66.3  & 63.5   & 52.3   \\
plant & sheep & sofa & train & tv  \\ \hline
21.9  & 42.2  & 48.1 & 59.8  & 47.1
\end{tabular}
}
\end{table}

Table \ref{table:FPSvsSize} clearly show that tiny-YOLO ROS for Pepper-Jetson TK1 achieves a near real-time inference speed of $\sim$4.8 FPS at an input image size of 320x320 while getting 47.69\% mAP on VOC 2007 test set. Using this same data, class-wise mAP is calculated on Table \ref{table:320classes}. This gives a detailed insight on the performance of the system at 320x320 resolution, where it is clear that correct detection of small objects like bottles or potted plants is a challenge for the system.  

\begin{figure}
\centering
\begin{subfigure}{.5\textwidth}
  \centering
  \includegraphics[width=0.975\linewidth]{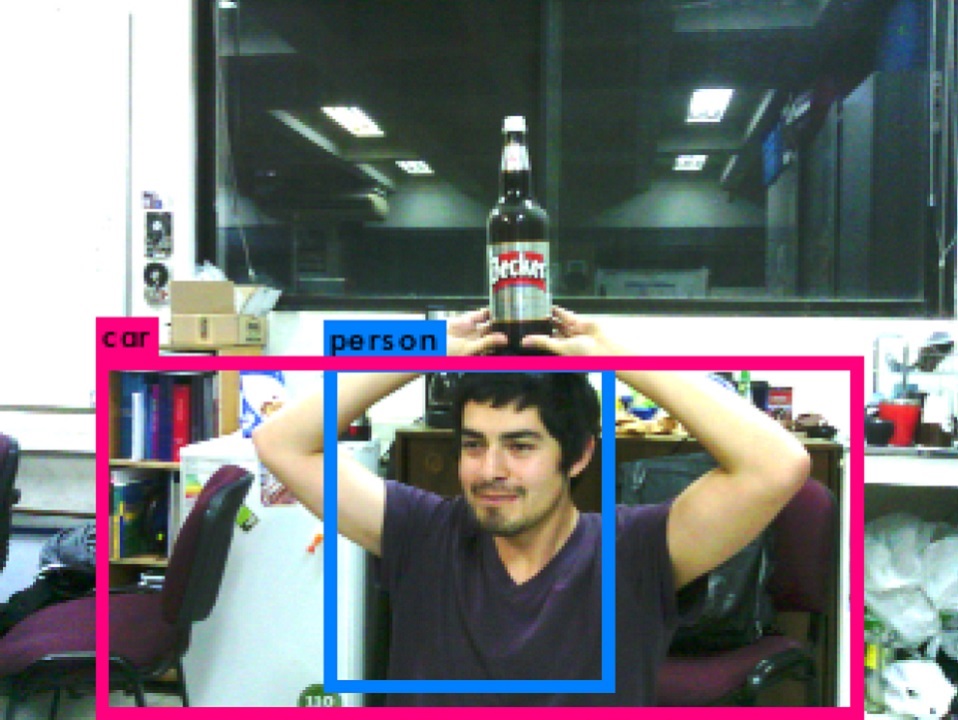}
  \caption{}
  \label{fig:y160}
\end{subfigure}%
\begin{subfigure}{.5\textwidth}
  \centering
  \includegraphics[width=0.975\linewidth]{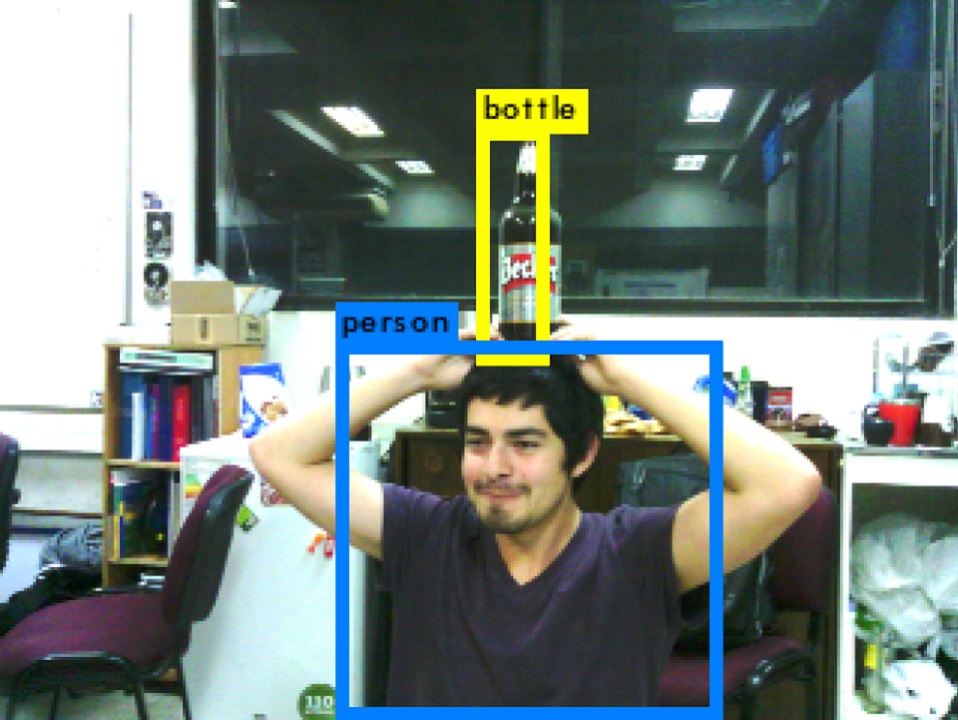}
  \caption{}
  \label{fig:y364}
\end{subfigure}
\caption{Detection examples of tiny-YOLO ROS for Pepper-Jetson TK1, at different resolution inputs. Color BB with labels shows detected objects. (a) Input image size of 160x160 pixels, where the model is unable to detect small objects like the bottle and mistakenly detect a car. (b) Input image size of 384x384 accurately detects multiple small to medium size objects.}
\label{fig:detectIm}
\end{figure}

After evaluating the system on classic data sets, we jumped to try it in an indoor environment, a place closer to the reality of what a service robot has to deal with. We analyzed different sizes of the input images, in order to select the one that allows obtaining an acceptable detection rate as well as processing speed. Fig. \ref{fig:y160} shows how for an image size of 160x160, the YOLO based detection system is unable to detect a bottle and it also throws false positives by detecting a non-existing car. On the other hand, when using a resolution of 384x384 pixels (Fig. \ref{fig:y364}) the YOLO based detection system is able to correctly detect the bottle and also better fits the BB for a person. 

\begin{figure}[!h]
  \begin{subfigure}[t]{.49\textwidth}
    \centering
    \includegraphics[width=\linewidth]{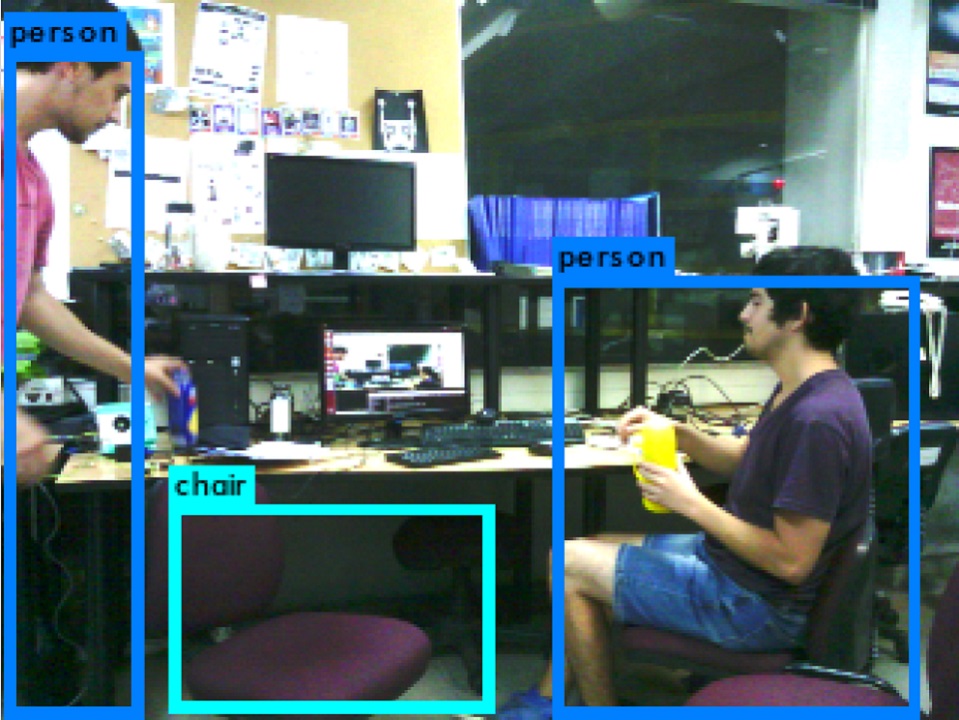}
    \caption{}
    \label{fig:multiObj320}
  \end{subfigure}
  \hfill
  \begin{subfigure}[t]{.49\textwidth}
    \centering
    \includegraphics[width=\linewidth]{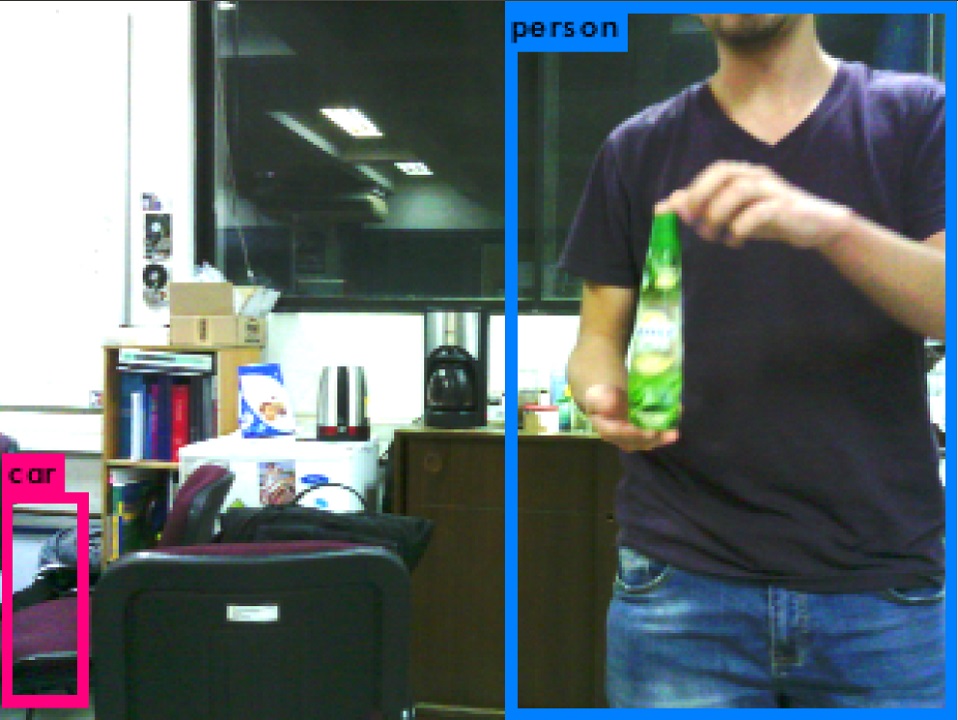}
    \caption{}
    \label{fig:falsePos320}
  \end{subfigure}

  \medskip

  \begin{subfigure}[t]{.49\textwidth}
    \centering
    \includegraphics[width=\linewidth]{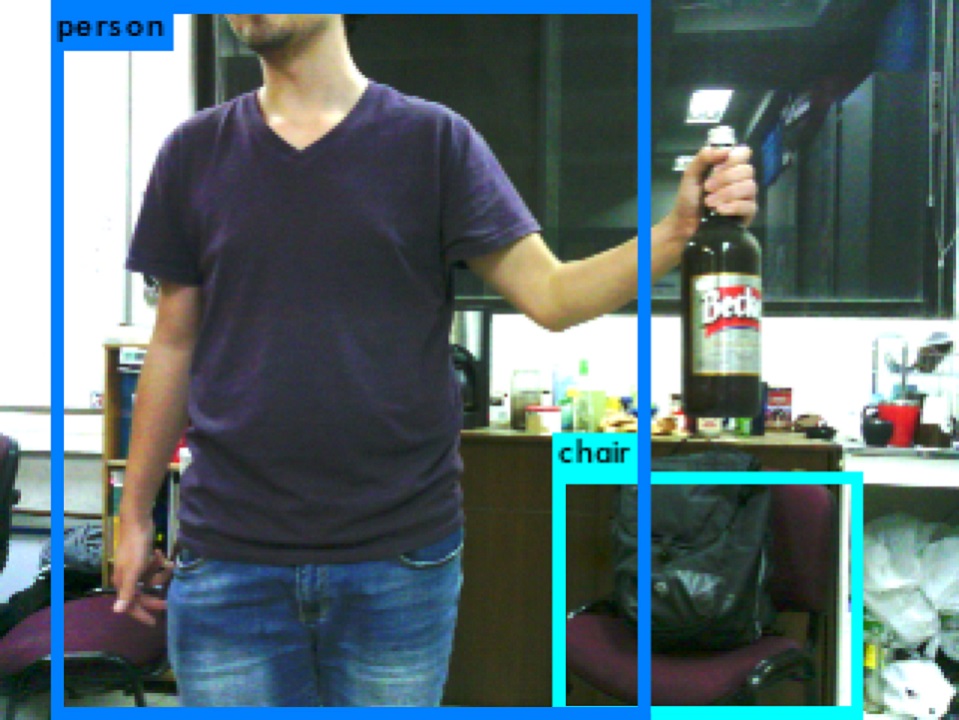}
    \caption{}
    \label{fig:noBottle320}
  \end{subfigure}
  \hfill
  \begin{subfigure}[t]{.49\textwidth}
    \centering
    \includegraphics[width=\linewidth]{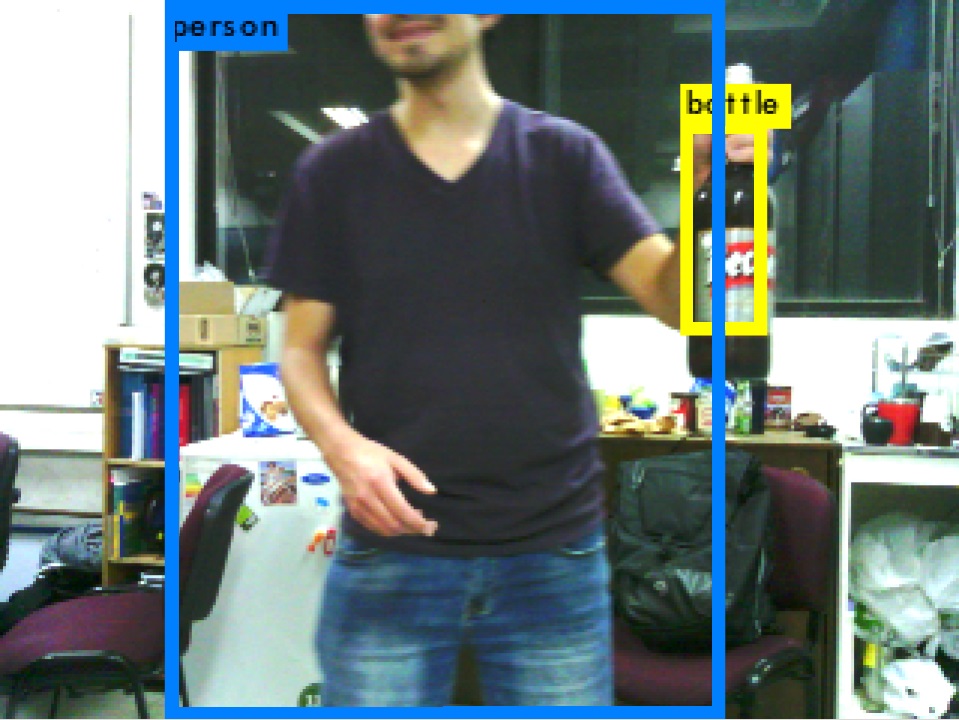}
    \caption{}
    \label{fig:bottle320}
  \end{subfigure}
  
\caption{Detection examples of tiny-YOLO ROS for Pepper-Jetson TK1, at $\sim$4.8 FPS and input image size of 320x320. This resolution allows detection of people and medium to large size objects, while struggles with small ones (a) Correct detection of multiple objects. (b) Example of an incorrect car detection, false positives are infrequent. (c) Correct detection of a person and a chair, but the bottle is ignored. (d) Correct detection of a small object when moving bottle from (c) closer to the camera.} 
\label{fig:detect320}
\end{figure}

When testing the detection system with input image of  320x320 pixels, it achieved near real-time inference speed, and it showed robust object detection of people and infrequent false positives. These results are shown in Fig \ref{fig:detect320}, where multiple sights of the indoor environment where shown to the system. Numerous object detection can be seen in Fig. \ref{fig:multiObj320} where 2 people and a chair are correctly detected. On the other hand, false positives that momentarily happen are depicted in Fig. \ref{fig:falsePos320}, by detection of a non-existing car. Small object detection difficulties can be seen in Fig. \ref{fig:noBottle320} and \ref{fig:bottle320}, where the bottle is not detected at first, but its recognized after moving it around, straighten it up and putting it closer to the camera of the Pepper robot. 

Therefore, we conclude that is very relevant to select an appropriate image size, and that an image size of 320x320 pixels allows obtaining a good tradeoff between detection accuracy and processing speed.

\section{Pepper Backpack}
Considering that using an external processing card (Jetson TK1) is the best alternative for implementing a YOLO based object detection system, we decided to design and built a backpack for Pepper, which can hold the card as well as its battery. 
\subsection{Mechanical Design}
The most important restriction to design the Pepper Backpack was to not modify the robot structure. For this reason, the attachment of the backpack to the robot should be non-invasive, therefore, an attachment using a suction cup is proposed.  

\begin{figure}[t]
\begin{subfigure}{.32\textwidth}
\centering
\includegraphics[width=1\linewidth]{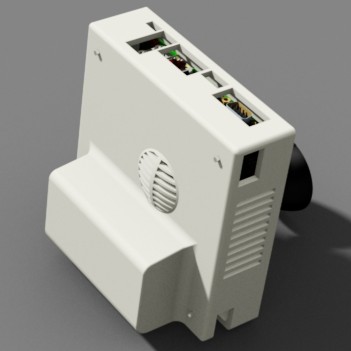}
\caption{}
\end{subfigure}\hfill
\begin{subfigure}{.32\textwidth}
\centering
\includegraphics[width=1\linewidth]{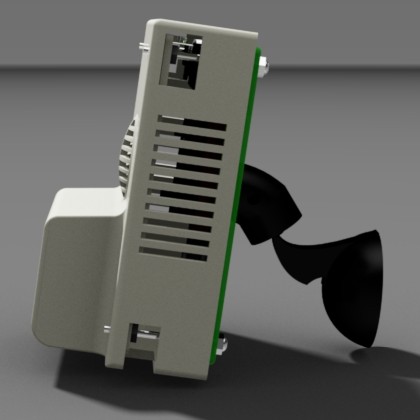}
\caption{}
\end{subfigure}\hfill
\begin{subfigure}{.32\textwidth}
\centering
\includegraphics[width=1\linewidth]{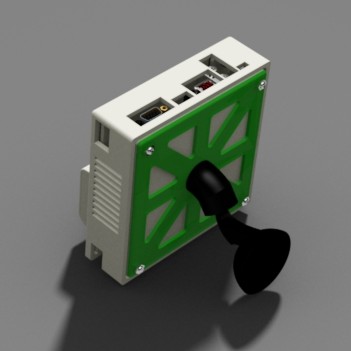}
\caption{}
\end{subfigure}
\caption{3D Renders of the Pepper Backpack. The main enclosure (white) is 3D printed. A commercial suction cup (black) is used to attach the backpack to Pepper. (a) Front view. (b) Lateral View. (c) Back view.}
\label{fig:3d-renders}
\end{figure}

\begin{figure}[h]
\centering
\begin{subfigure}{.45\textwidth}
  \centering
  \includegraphics[width=0.975\linewidth]{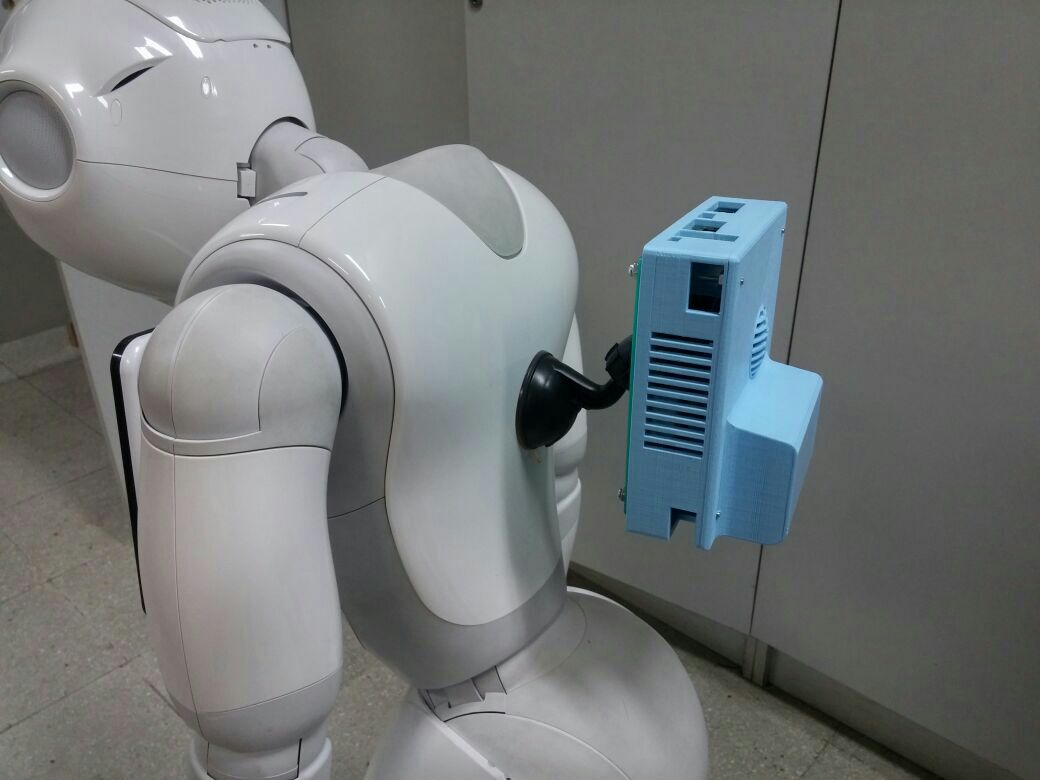}
  \caption{}
  \label{fig:pepperWhole}
\end{subfigure}%
\begin{subfigure}{.45\textwidth}
  \centering
  \includegraphics[width=0.975\linewidth]{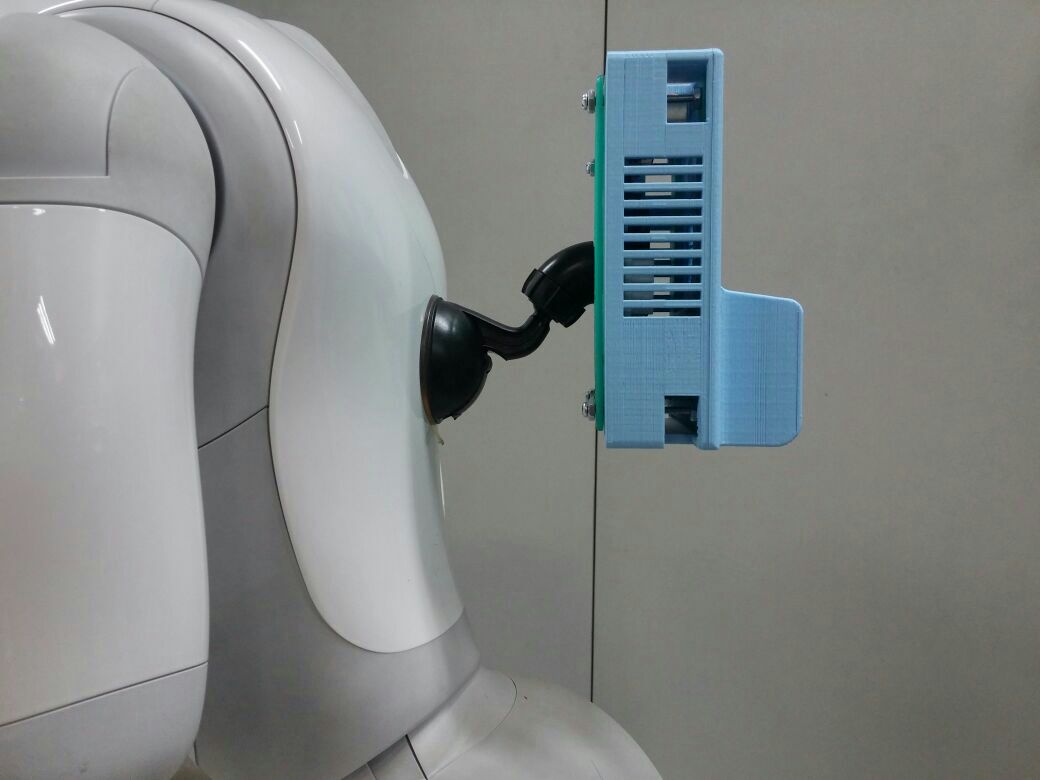}
  \caption{}
  \label{fig:pepperLat}
\end{subfigure}
\caption{Pepper robot with the Jetson TK1 backpack attached to its lower back. (a) Perspective view. (b) Lateral close-up tho backpack.}
\label{fig:backpack-photos}
\end{figure}

The main enclosure of the Jetson computer corresponds to one acrylic plate in the base and a 3D printed case with a battery compartment mounted on the front of the board. A custom joint was 3D printed to connect the enclosure of the Jetson and the suction cup structure. Whole backpack renders can be seen in Fig. \ref{fig:3d-renders}, while perspectives of the backpack attached to Pepper robot through a 5.8 $mm$ of diameter commercial suction cup can be seen in Fig. \ref{fig:backpack-photos}.

The CAD models, list of components and materials required to build the backpack can be found in the Supplementary Material section.

\subsection{Hardware}
\subsubsection{Nvidia Jetson TK1}
The Nvidia Jetson TK1 main processor is a Nvidia Tegra K1 which is a CPU+GPU+ISP single chip.  The existence of a GPU in the Jetson allows it to run state-of-the-art Deep Learning algorithms.

The Nvidia Jetson TK1 comes with a power supply that can provide 12 $volts$ and 5 $ampere$ maximum. These requirements are fulfilled by a 3-Cell LiPo battery. The battery is connected to the Jetson through a standard 2.1x5.5$mm$ barrel jack. The typical power-draw of the Jetson does not surpass 10 $watts$ \cite{jetsonpowerdraw}. It is important to note that Pepper does not provide a power output, thus using it as a source of power for the Jetson card is not possible without structural modifications of Pepper.

To connect the Jetson computer to Pepper, and enable data stream between platforms, the Ethernet Gigabit port of both is used.

\section{Conclusions and Future Work}
In this work, we studied the use of YOLO on the Pepper's computer, by testing different versions of the model and variation of their parameters. We concluded that it is not possible to surpass the 1 FPS minimum limit in order to achieve near real-time object detection. As an alternative, we introduced external image processing, choosing the Jetson TK1 computer as the device to run tiny-YOLO at an input image resolution of 320x320, which demonstrated to be the best-suited model to reach high-speed processing of $\sim$4.8 FPS. By using a smaller input image size we gained speed, at cost of performance, this was reflected on an mAP of 47.69\% over VOC 2007 test set, and a low capacity of detection for small objects on real-world indoor environments.  

To enable high-speed communication between Pepper and the Jetson TK1 computer we use an Ethernet Gigabit connection.  More important is the fact that we directly attached the board onto Pepper through a custom made backpack, which does not affect the movement of the robot.

As a future work task, we propose usage of another external computer besides Jetson TK1, because it may not be the most suitable platform to best exploit inference speed of tiny-YOLO. This statement lies in the fact that the Jetson TK1 is a 32-bit system that supports up to CUDA 6.5 version, which does not allow usage of the deep neural networks dedicated library cuDNN. Platforms such as Jetson TX1 or Jetson TX2 which allow higher versions of CUDA and thus cuDNN usage may outperform results presented in this work.

Finally, all the necessary files to replicate the project will be publicly available.

\section*{Supplementary Material}
All the necessary to replicate the project resides in two GitHub repositories. One repository provides the CAD models and list of components to build the Pepper Backpack: \href{https://github.com/uchile-robotics/pepper-backpack}{https://github.com/uchile-robotics/pepper-backpack}. The other repository corresponds to a fork of \texttt{darknet\_ros} with instructions to run tiny-YOLO ROS on the Pepper-Jetson TK1: \href{https://github.com/uchile-robotics-forks/darknet\_ros}{https://github.com/uchile-robotics-forks/darknet\_ros}

\section*{Acknowledgements}
This research was partially funded by FONDECYT Project 1161500.

\bibliographystyle{unsrt}
\bibliography{bibliografia}
\end{document}